# Motion deblurring of faces

Grigorios G. Chrysos · Paolo Favaro · Stefanos Zafeiriou

**Abstract** Face analysis is a core part of computer vision, in which remarkable progress has been observed in the past decades. Current methods achieve recognition and tracking with invariance to fundamental modes of variation such as illumination, 3D pose, expressions. Notwithstanding, a much less standing mode of variation is motion deblurring, which however presents substantial challenges in face analysis. Recent approaches either make oversimplifying assumptions, e.g. in cases of joint optimization with other tasks, or fail to preserve the highly structured shape/identity information. Therefore, we propose a data-driven method that encourages identity preservation. The proposed model includes two parallel streams (sub-networks): the first deblurs the image, the second implicitly extracts and projects the identity of both the sharp and the blurred image in similar subspaces. We devise a method for creating realistic motion blur by averaging a variable number of frames to train our model. The averaged images originate from a $2MF^2$ dataset with 10 million facial frames, which we introduce for the task. Considering deblurring as an intermediate step, we utilize the deblurred outputs to conduct a thorough experimentation on high-level face analysis tasks, i.e. landmark localization and face verification. The experimental evaluation demonstrates the superiority of our method.

**Keywords** Learning motion deblurring · face deblurring

G. Chrysos · S. Zafeiriou
Department of Computing, Imperial College London, 180 Queen's Gate, London SW7 2AZ, UK
E-mail: {g.chrysos, s.zafeiriou}@imperial.ac.uk
P. Favaro
Department of Informatics, University of Bern, Neubruckstrasse 10, CH-3012 Bern, Switzerland
E-mail: paolo.favaro@inf.unibe.ch

## 1 Introduction

Defocus and motion blur are the two most common types of blur that appear in images[1]. Motion blur is caused by a change of the relative position of the camera-scene system during the sensor exposure. This image degradation is caused by i) camera shake, ii) object movement, iii) a combination of the two. This relative movement is typically unknown to the method (blind deblurring). The algorithm is given just a blurry image and needs to recover the corresponding sharp one. Deblurring is significant for both high-end systems (e.g. human vision) and for computational tasks. For instance, an emerging trend in face analysis is to learn blur-invariant representations. To that end, we introduce a method for deblurring facial images which suffer from (severe) motion blur.

Frequently, face analysis tasks involve (implicitly) learning a low-dimensional space with the desired properties. This low-dimensional space is invariant to modes of variations influencing the performance of the task, e.g. rotation-invariant face recognition. Such invariance has been largely achieved for fundamental modes of variation, e.g. rotation, illumination (Tran et al (2017)). Recent works for approaching blur invariance have emerged, e.g. Ding and Tao (2017), however invariance in this mode is far from accomplished yet. This can be partly attributed to the underdetermined nature of the task; there are infinite combinations for sharp images, blur models and non-linear functions in the images process. Despite that, we are interested only in the images that span the manifold of the natural images; deblurring

---

[1] Defocus has significant applications, however it can be simulated with less complicated kernels than motion blur, e.g. in Ding and Tao (2017). Thus we focus exclusively on motion blur in this work.

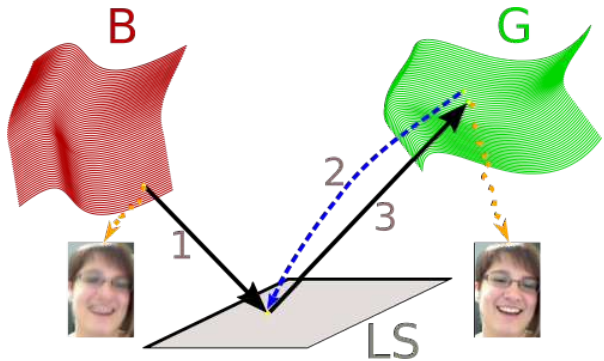

Fig. 1: The recent approaches in deblurring, e.g. Kupyn et al (2017), project an image from the manifold of blurry images **B** to a latent space **LS** and then to the manifold of sharp images **G**, i.e. they use steps 1 and 3. In our approach we insert step 2, i.e. during training time we project from **G** to the latent space as an implicit identity preservation. A pair of blurry/sharp images, sampled from the **B** and **G** manifolds, are depicted (dashed orange line from the respective manifold). All figures in this paper are best viewed in color.

methods capitalize on this restriction with various ways, e.g. priors or domain-specific knowledge.

Deblurring an object can be tackled with two types of methods: i) generic object deblurring, ii) domain-specific methods. Even though the task of generic object deblurring is well-studied (Tekalp et al (1986); Cho and Don (1991); Levin et al (2009); Pan et al (2014a); Nah et al (2017)), generic methods yield suboptimal results on faces. The reason is that generic methods typically rely on gradient-based information which cause them to fail in an object with many flat regions, e.g. human face. Moreover, the face includes highly structured shape, which is not utilized by the priors of generic object deblurring. Therefore, domain-specific methods could offer a better alternative for deblurring faces.

The domain-specific methods in face deblurring can be classified in two categories: i) those that include joint optimization for solving another task (Liao et al (2016); Nguyen et al (2015); Ding and Tao (2017)), ii) those that utilize the geometric information, e.g. shape/contour (Pan et al (2014a)). The former methods optimize over a face analysis task, e.g. face recognition, and either include deblurring related priors in their cost function or add blurry examples in their training set, i.e. implicitly learn blur-invariant representations. However, joint optimization methods use oversimplifying assumptions for blur to enable easier convergence of the tasks. The latter methods utilize the contour or a sparse shape of the face to guide their optimization.

Geometric information lead to successful outcomes for synthetic/mild blurs, however extracting geometric cue information for real-world blurry images is not trivial.

In our work we restrict further the solution space by realizing that our deblurred images should belong in the manifold shaped by images of that identity. We introduce a method that performs (implicit) identity preservation of the face. Preserving identity information is significant for various face analysis tasks (Ding and Tao (2017)). In our case, we design an architecture that includes two parallel streams (during training); the first one accepts the blurry image, while the second extracts facial representations (similar to an auto-encoder). By sharing the weights in part of the two streams, the first stream is encouraged to have representations similar to those of the second stream. To ensure that our network performs well even under severe blur, we perform a two-step deblurring, where the first step includes a strong discriminative network to restore the low frequencies and then restore the high frequency details with a network that provides high fidelity information. As a supervised data-driven method, our method requires pairs of blurry and ground-truth (sharp) images for training.

Collecting pairs of blurry/sharp images for training a deblurring network is a laborious and expensive process (it requires specialized hardware). Such databases, e.g. Su et al (2017); Nah et al (2017); Kim et al (2017); Noroozi et al (2017), include spatial and temporal limitations, i.e. capturing covers only a specific time-span and is restricted geographically. In addition, due to the specialized equipment such methods cannot capitalize on the vast amount of data available online. This impediment is tackled by i) convolving sharp images with synthetic blur kernels, ii) simulating motion blur by averaging sharp frames. Synthetic blurs (Hradiš et al (2015)) can only model constrained conditions, while generalization to real-world blurry images remains questionable (Lai et al (2016)). Moreover, in our domain synthetic blurs cannot capture natural facial movement (or deformations). On the contrary, averaging sharp frames creates a natural movement. If the camera includes a high fps[2], then by averaging sequential frames, we simulate the natural displacement of the scene objects. The naive solution is to average a predefined number of frames, which understandably is quite restrictive. For instance, a sudden camera shake/zoom-in causes an excessive blur that is not present in the rest of the frames. To ameliorate that, we consider the simulation of blurry frames as a multivariate function that depends on the number of frames averaged, the overlap of fiducial points, the optical flow and the image quality.

---

[2] Most commercial cameras capture at least 30 fps, which is reasonably fast.

A direct drawback of the averaging scheme is the requirement for vast amount of frames. The lack of such facial data is partially the reason why face deblurring is understudied. To that end, we introduce $2MF^2$, a dataset with 4,000 videos which accumulate to 10 million frames. We use $2MF^2$ videos to generate blurry images and then train our system. $2MF^2$ includes the largest number of long videos of faces (each video has a minimum of 500 frames), while it includes multiple videos of the same identity in different scenes and conditions.

Following recent trends in deblurring (Hradiš et al (2015); Kupyn et al (2017)), we consider deblurring as an intermediate step in face analysis tasks. Thus, apart from the typical image quality metrics, we perform a thorough experimentation by utilizing the deblurred outcomes for two different tasks. Namely, we perform landmark localization, and face verification on two different datasets. Both tasks have solid quantitative figures of merit and can be used to evaluate the final result, hence implicitly verify whether the deblurred images indeed resemble the samples from the distribution of sharp images.

This work is an extension of Chrysos and Zafeiriou (2017b), where we introduced the first network used for deblurring facial images. The current work has substantial extensions. First of all, in the original work there was no explicit effort to preserve the facial identity. In addition, the sparse shape utilized for the original work did not work well in the general case; we allow the network to learn the shape implicitly. The architecture has been redesigned from scratch; the ResNet of He et al (2016) in the previous version is much different than the customized architecture we devise in this work to facilitate identity preservation. Last but not least, the experimental section has been completely redesigned; in this work we not only use standard quality metric for deblurring, but also utilize the deblurred images as an intermediate step for experimenting with higher level tasks in face analysis.

Our contributions can be summarized as:

- We introduce the first learning-based architecture for deblurring facial images. To train that network, we introduce a new way to simulate motion blur from existing videos.
- We introduce $2MF^2$ dataset that includes over 10 million frames; the frames are utilized for simulating motion blur.
- We conduct a thorough experimental evaluation i) with image quality metrics and ii) by utilizing the deblurred images in other tasks. The deblurred images are compared in sparse regression and classification in face analysis tasks. Our comparisons involve deblurring over 60,000 images for each method, which consists one of the largest testsets used for comparing deblurring methods.

We consider our proposed method as a valuable addition to the research community, hence the blurry/sharp pairs along with the frames of $2MF^2$ will be released upon the acceptance of the paper.

**Notation:** A small (capital) bold letter represents a vector (matrix); a plain letter designates a scalar number. Table 1 describes the primary symbols used in the manuscript.

Table 1: Summary of primary symbols. '#' abbreviates the 'number of'.

| Symbol | Dimen. | Definition |
|---|---|---|
| $\boldsymbol{I}_s$ | $\mathbb{R}^{h \times w}$ | Latent sharp image. |
| $\boldsymbol{I}_{bl}$ | $\mathbb{R}^{h \times w}$ | Blurry image. |
| $L$ | $\mathbb{R}$ | # frames averaged (motion blur). |
| $\lambda_{\{ci, cg, p, r\}}$ | $\mathbb{R}$ | Regularization hyper-parameters. |

## 2 Related work

We initially provide an overview of the recent advances with Generative Adversarial Networks (core component of our system), then recap the literature on deblurring and sequentially study how blur is studied in face analysis tasks.

### 2.1 Generative Adversarial Network

Generative Adversarial Networks (GANs) by Goodfellow et al (2014) have received wide attention. GANs sample noise from a predefined distribution (e.g. Gaussian) and learn a mapping with a signal from the domain of interest. Several extensions have emerged, like using convolutional layers instead of fully connected in Radford et al (2015), feeding a (Laplacian) pyramid for coarse-to-fine generation in Denton et al (2015), jointly training a GAN with an inference network in Dumoulin et al (2016), learning hierarchical representations in Huang et al (2017). Alternative cost functions and divergence metrics have been proposed (Nowozin et al (2016); Arjovsky et al (2017); Mao et al (2017)). In addition, several approaches for improving the training of GAN's have appeared (Salimans et al (2016); Dosovitskiy and Brox (2016)). GANs have been used for unsupervised (Radford et al (2015); Arjovsky et al (2017)), semi-supervised (Odena (2016)) and supervised learning (Mirza and Osindero (2014); Ledig et al (2017); Isola et al (2017); Tulyakov et al (2017)). The proliferation of the works with GANs can be attributed to their ability to preserve high texture details and model highly complex distributions.

## 2.2 Deblurring

Deblurring defines the computational task of reversing the unknown blur that has been inflicted to a sharp image $I_s$. In the previous few decades, the problem was formulated as an energy minimization with heuristically defined priors, which reflect image-based statistics or domain-specific knowledge. However, aside of the computational cost of these optimization methods (typically they require over a minute for deblurring an image of 300x300 resolution), their prior consists their Achilles heel. Deep learning methods alleviate that by learning from data.

**Energy optimization methods**: The blurry image $I_{bl}$ is assumed as the convolution of the latent sharp image $I_s$ and a (uniform) kernel $K$, mathematically expressed as $I_{bl} = I_s * K + \epsilon$, where $\epsilon$ denotes the noise. Deblurring is then formulated as minimization of the cost function

$$I_s = \arg\min_{\tilde{I}_s}(||I_{bl} - \tilde{I}_s * K||_2^2 + f(I_{bl}, K)). \quad (1)$$

with $f(I_{bl}, K)$ a set of priors based on generic image statistics or domain-specific priors. These methods are applied in a coarse-to-fine manner; they estimate the kernel and then perform non-blind deconvolution.

The blur kernel $K$ and the latent image $I_s$ are estimated in an alternating manner, which might lead to a blurry result if a joint MAP optimization is performed (Levin et al (2009)). They suggest instead to solve a MAP on the kernel with a gradient-based prior that reflects natural image statistics. Pan et al (2014b) apply an $\ell_0$ norm as a sparse prior on both the intensity values and the image gradient for deblurring text. Hacohen et al (2013) support that the gradient-based prior alone is not sufficient. They introduce instead a prior that locates dense correspondences between the blurry image and a similar sharp image, while they iteratively refine the correspondence estimation, the kernel and the sharp image estimation. Their core idea relies on the existence of a similar reference image, which is not always available. A generalization of Hacohen et al (2013) is the work of Pan et al (2014a), which relaxes the existence of a similar image with an exemplar dataset. The assumption is that there is an image with a similar contour in the exemplar dataset. However, the contour of an unconstrained object or the similarities between contours are not trivially found, hence Pan et al (2014a) restrict the task to face deblurring to profit from the constrained shape structure. At test time, a search in the dataset with the exemplar images is performed; the exemplar image with a contour similar to the test image is then used to initialize the blind estimation iterations. Pan et al (2014a) demonstrate how this leads to an improved performance. Unfortunately, the noisy contour matching process along with the obligatory presence of a similar contour in the dataset limit the applications of this work. Huang et al (2015) recognize the deficiencies of this approach and propose to perform landmark localization before extracting the contour. They effectively replace the exemplar dataset matching by training a localization technique with blurry images, however their approach still suffers in more complex cases (they use a few synthetic kernels) or even more in severe blur cases.

In contrast to the gradient-based priors, Pan et al (2016) introduce a prior based on the sparsity of the dark channel. The dark channel is defined as the pixel with the lowest intensity in a spatial neighborhood. Pan et al (2016) prove that the intensity of the dark channel is increased from the blurring process; they demonstrate how the sparsity of the dark channel leads to improved results.

Even though the aforementioned methods provably minimize the energy of a blurred image, their strong assumptions (e.g. non-informative, hard-coded priors) consist them both computationally inefficient[3] and with poor generalization to real-world blurry images (Lai et al (2016)).

**Learning-based methods for motion blur removal**: The experimental superiority of neural networks as function approximators have fuelled the proliferation of deblurring methods learned from data. The blurring process includes several non-linearities, e.g. the camera response function, lens saturation, depth variation, which the aforementioned optimization methods cannot handle. Conversely, neural networks can approximate these non-linearities and learn how to reverse the blur by feeding them pairs of sharp and blurry images.

There are two dominant approaches: i) use a data-driven method to learn an estimate; then refine the kernel/image estimation with classic methods (Sun et al (2015); Chakrabarti (2016)), ii) let the network explicitly model the whole process and obtain the deblurred result (Hradiš et al (2015)).

In the former approach, Schuler et al (2016) design a network that imitates the optimization-based methods, i.e. it iteratively extracts features, estimates the kernel and the sharp image. Sun et al (2015) learn a convo-lutional neural network (CNN) to recognize few pre-defined motion kernels and then perform a non-blind deconvolution. Chakrabarti (2016) proposes a patch-based network that estimates the frequency information for uniform motion blur removal. Gong et al (2017) train a network to estimate the motion flow (hence the per-pixel blur) and then perform non-blind deconvolution.

---

[3] The coarse-to-fine procedure is executed in a loop hundreds or even thousands of times to return a deblurred image; as indicated by Chakrabarti (2016) some of them might even require hours for deblurring a single image.

The second approach, i.e. modelling the whole process with a network, is increasingly used due to the increased capacity of the networks. Noroozi et al (2017); Nah et al (2017) introduce multi-scale CNNs and learn an end-to-end approach where they feed the blurry image and obtain the deblurred outcome. Nah et al (2017) also include an adversarial loss in their loss function. A number of very recent works utilize adversarial learning to learn an end-to-end mapping between blurry and sharp images (Ramakrishnan et al (2017); Kupyn et al (2017)).

The works utilizing adversarial learning are the closest to our methods, however there a number of significant differences in our case. First of all, we create an architecture designed to preserve the identity of the object; we also approach the task as a two step process where in the first step we restore the low frequency components and then refine the high frequency details by adversarial learning.

### 2.3 Face analysis under blur

As face analysis consists a core application of computer vision, the need for studying the blurring process is increasingly emphasized, e.g. in Zafeiriou et al (2017). The face detector of Liao et al (2016) introduces the NPD features and experimentally verify how robust they are under different blur conditions. Estimating the age from a blurry face is the task of Nguyen et al (2015), who introduce an optimization method that estimates the motion blur; classify the blurry image based on the blur and deblur accordingly based on the category.

Facial blur poses a major challenge in face recognition, hence few efforts explicitly include blurry images in the learning process. Nishiyama et al (2011) construct a identity-invariant feature space, which has the property that images degraded by similar blur are clustered together. Then, cluster-based deblurring is performed. Even though the idea works well for synthetic images, the clusters of real-world blurs are less distinct. Gopalan et al (2012) derive a blur-robust descriptor for face recognition, based on simplifying assumptions about i) the convolution with a specific size kernel, ii) no-noise case. Ding and Tao (2017) learn blur-invariant representations by feeding simultaneously a blurry and a sharp image to the network.

## 3 Method

In this section we outline the concept of conditional GAN (cGAN), then we introduce our proposed architecture and elaborate how to generate realistic training pairs for our application.

### 3.1 conditional GAN

GAN consists of two parts, a generator $G$ and a discriminator $D$; the generator samples from a predefined distribution and tries to model the true distribution of the data $p_d$. The discriminator tries to distinguish between real (sampled from the true distribution) and fake signals (sampled from the model's distribution). Conditional GAN (cGAN) by Mirza and Osindero (2014) extends the formulation by conditioning the distributions with additional labels. If $p_z$ denotes the distribution of the noise, $s$ the conditioning label and $y$ the data, the objective function is expressed as:

$$\mathcal{L}_{cGAN}(G, D) = \mathbb{E}_{s,y \sim p_d(s,y)}[\log D(s, y)] + \\ \mathbb{E}_{s \sim p_d(s), z \sim p_z(z)}[\log(1 - D(s, G(s, z)))] \quad (2)$$

This objective function is optimized in an iterative manner, as

$$\min_{w_G} \max_{w_D} \mathcal{L}_{cGAN}(G, D) = \mathbb{E}_{s,y \sim p_d(s,y)}[\log D(s, y; w_D)] + \\ \mathbb{E}_{s \sim p_d(s), z \sim p_z(z)}[\log(1 - D(s, G(s, z; w_G); w_D))]$$

where $w_G, w_D$ denote the generator's and the discriminator's parameters respectively.

In our case the output domain (deblurred results) are dependent on the input domain (blurred input); we utilize a cGAN that is precisely defined for such a task. Similarly to our case, conditional GANs have been applied to diverse image processing tasks, since they output photo-realistic images. Recent applications include photo-realistic image synthesis by Ledig et al (2017), style transfer by Yoo et al (2016), inpainting by Pathak et al (2016), image-to-image mappings by Isola et al (2017), video generation by decoupling content from motion in Tulyakov et al (2017), image hallucination in Xu et al (2017).

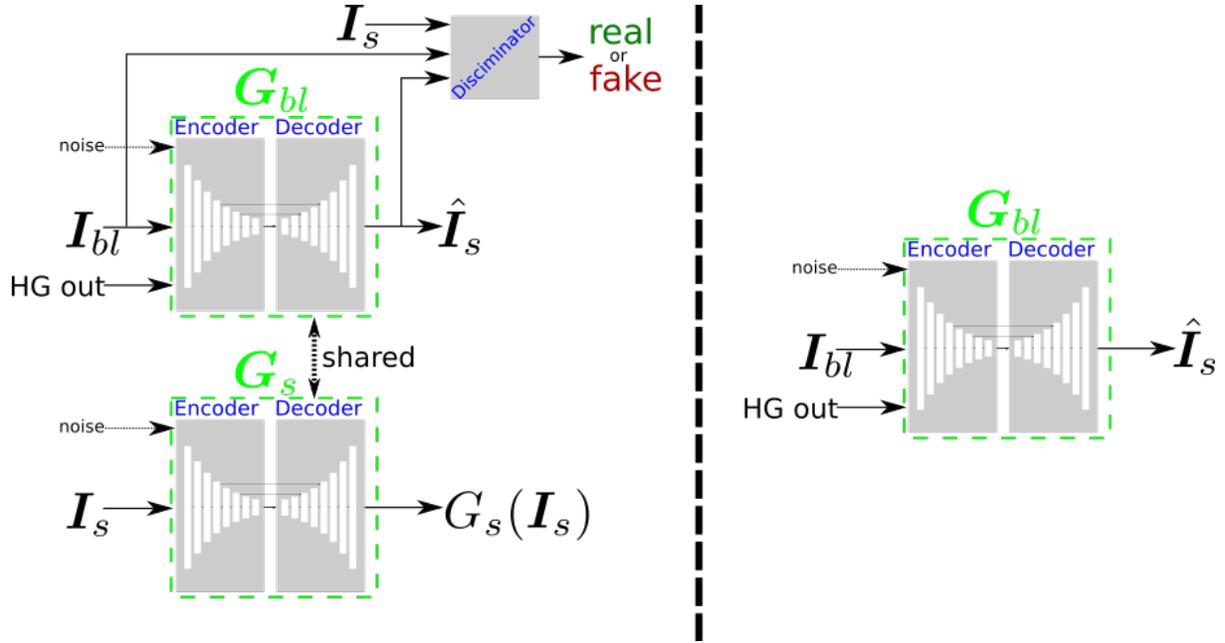

Fig. 2: Model architecture. The architecture during training time is depicted in the left. 'HG out' signifies the additional label used as input, i.e. the output of the HG network, while the weight sharing of the two decoders is denoted with the dashed line. During prediction (testing time), the network is greatly simplified; it is depicted in the right part of the figure.

## 3.2 Model architecture

Even though components of our network's architecture are based on a conditional GAN, the two fundamental modules are i) the identity preservation module, ii) domain-specific module for deblurring. These two modules can be used as plug-in parts in several architectures, i.e. creating a parallel stream for identity preservation or the two step process for gradual restoration.

We achieve the preservation of identity by including two parallel streams, i.e. sub-networks; the first deblurs the image, the second implicitly extracts and projects the identity of both the sharp and the blurred image in similar subspaces. The two streams are two similar generators implemented in an encoder-decoder style. The first generator, denoted as $G_{bl}$, accepts the blurred images, while the second, denoted as $G_s$, accepts the ground-truth images. The decoders of the two copies have shared weights, since we want the latent representations of the sharp and blurry images to be similar. In addition to the two generators, we add a discriminator, which along with $G_{bl}$ form a cGAN.

The domain-specific change is based on the underdetermined nature of deblurring. Reversing the blurring process requires several convolutional layers; more than the few layers a conditional GAN typically includes. To remedy that, we create a two step process; the output of the first step is used as an additional label in the $G_{bl}$. Specifically, we add the strong-performing hourglass network (HG) by Newell et al (2016) as a first step; the output of the hourglass network is the new label which we condition $G_{bl}$. HG has a top-down approach and combines low-level features from different resolutions. HG unscrambles the low-frequency details of the image and then the cGAN can generate the high-frequency details to obtain the final outcome.

The complete architecture is visualized in Fig. 2; the implementation details are analyzed in sec. 5. We design a loss function with four terms. Aside of the adversarial loss $\mathcal{L}_{cGAN}(G_{bl}, D)$, which is computed based on the $G_{bl}$ generator's output, we add a content loss, a projection loss and a reconstruction loss.

The content loss consists of two terms that compute the per-pixel difference between the generator's output ($G_{bl}(\boldsymbol{I}_{bl})$) and the sharp (ground-truth) image. The two terms are i) the $\ell_1$ loss between the ground-truth image and the output of the generator, ii) the $\ell_1$ of their gradients; mathematically expressed as:

$$\mathcal{L}_c = \lambda_{ci}||G_{bl}(\boldsymbol{I}_{bl}) - \boldsymbol{I}_s||_{\ell_1} + \lambda_{cg}||\nabla G_{bl}(\boldsymbol{I}_{bl}) - \nabla \boldsymbol{I}_s||_{\ell_1} \quad (3)$$

where $\lambda_{ci}, \lambda_{cg}$ are two hyper-parameters.

The projection loss[4] enables the network to match the data and the model's distribution faster. The intuition is that to match the high-dimensional distribution of the data with the model one, we can encourage their projections in lower-dimensional spaces to be similar. To avoid adding extra parameters or designing a hard-coded projection, we utilize the projection of the discriminator. If $\pi$ denotes the projected features from the penultimate layer of the discriminator, then:

$$\mathcal{L}_p = ||\pi(G_{bl}(\boldsymbol{I}_{bl})) - \pi(\boldsymbol{I}_s)||_{\ell_1} \qquad (4)$$

Last but not least, the reconstruction loss ensures that $G_s$ captures the identity of the person. We penalize any dissimilarities between the reconstructed image from the identity preservation stream, i.e.

$$\mathcal{L}_r = ||G_s(\boldsymbol{I}_s) - \boldsymbol{I}_s||_{\ell_1} \qquad (5)$$

The total loss function is expressed as:

$$\mathcal{L} = \mathcal{L}_{cGAN} + \mathcal{L}_c + \lambda_p \mathcal{L}_p + \lambda_r \mathcal{L}_r \qquad (6)$$

where $\lambda_p, \lambda_r$ are hyper-parameters.

### 3.3 Training data

Pairs of sharp images along with their corresponding motion blurred images are required to learn the model. To understand how to create such pairs, we examine briefly how the motion blur is generated. To capture an image, the aperture is opened, accumulates light from the dynamic scene and then creates the integrated result. The process can be formulated as an integration of all the light accumulated, or for the discrete machines the sum of all values as following:

$$\boldsymbol{I}_{bl} = \psi(\int_0^T \boldsymbol{I}_s(t)dt) \approx \frac{1}{K+1}\psi(\sum_{k=0}^K \boldsymbol{I}_s[k]) \qquad (7)$$

where $\boldsymbol{I}_{bl}$ denotes the blurry image, $\boldsymbol{I}_s$ the latent sharp image, $T, K$ the duration in continuous/discrete time that the aperture remains open. The function $\psi$ expresses the unknown non-linear relationships that the imaging process includes, for instance lens saturation, sensor sensitivity. We cannot analytically compute the function; we can only approximate it, but this remains challenging as studied by Grossberg and Nayar (2003); Tai et al (2013).

The aforementioned blurring process can be approximated computationally. The three dominant approaches for creating pairs of sharp/blurry images are the following: a) use specialized equipment/hardware setup, b) simulate motion blur by averaging frames, c) convolve image with synthetic kernels. Synthetic motion blur was created by kernels in the past, please find a thorough review in Lai et al (2016). Such synthetic blurs assume a static scene and are only applied in the 2D image plane, which consists them very simplistic. In addition, deblurring synthetically blurred images does not have a high correlation with deblurring on unconstrained conditions (Lai et al (2016)).

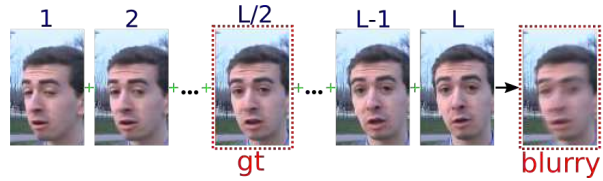

Fig. 3: Averaging scheme to simulate the motion blur. Averaging $L$ frames generates a realistic motion blur and we can consider the middle frame ($L/2$) as the gt.

Using specialized equipment has appeared in the recent works of Su et al (2017); Nah et al (2017); Kim et al (2017); they utilized GoPro Hero cameras to capture videos at 240 fps. The frames are post-processed to remove the blurry ones; sequential frames are then averaged to simulate the motion blur. The main benefit of such high-end devices is that they can reduce the amount of motion blur per frame; virtually ensuring that there will be minimal relative motion between the 3D scene and the camera at each capture. However, the spatial and temporal constraints of such a collection consist a major limitation. Even though a significant effort to capture diverse scenes might be required, the collection spans small variety of scenes with a constrained time duration (limited number of samples). Additionally, only specific high-fps capturing devices can be used and even those under restricted conditions, e.g. good lighting conditions.

On the other hand, the simulation entails averaging sequential frames of videos captured from commodity cameras (30 - 60 fps). Such videos are abundant in online platforms, for instance in YouTube hundreds of hours of new content are uploaded per minute. Content covers both short amateur clips and professional videos, while the scenes vary from outdoor clips in extreme capturing conditions to controlled office conditions. Previously to our work, Wieschollek et al (2017) utilize such sources to simulate motion blur. For each pair of sequential frames the authors utilize a bidirectional optical flow to generate a number of sub-frames,

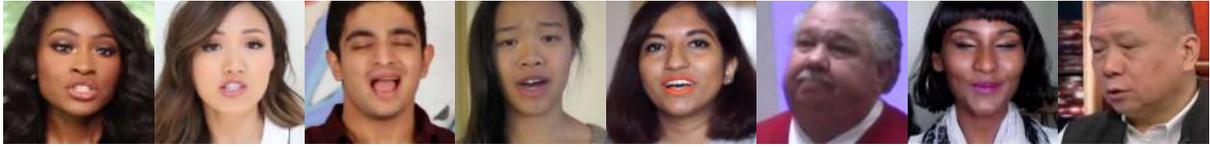

Fig. 4: Visualization of $2MF^2$ dataset samples.

which are then averaged to generate the blurry-sharp pair.

In our case, we utilize videos captured with commodity cameras and are available in the internet. We do not resort to any frame warping as this might lead to artifacts not present in the real-world motion blur cases. In lieu, we average frames from the training videos, more formally the precise blur is computed as:

$$\tilde{\mathbf{I}}_{bl} = \frac{1}{L} \sum_{l=-\frac{L-1}{2}}^{\frac{L-1}{2}} \hat{\psi}(\mathbf{I}_s[l]) \tag{8}$$

where $L$ denotes the number of frames in the moving average, $\hat{\psi}$ a function that is learned and approximates $\psi$.

The number of frames summed, i.e. $L$, varies and depends on the cumulative relative displacement of the face/camera positions. The number $L$ is dynamically decided based on the current frames; effectively we generate a sub-sequence of $L$ frames, we average the intensities of those to obtain the blurry image and consider the middle as the ground-truth image; the process is illustrated in Fig. 3. We continue adding frames to the sub-sequence until stop conditions are met. The conditions that affect the choice of $L$ are the following: i) there should be an overlap of some of the facial semantic parts between frames, ii) the quality of the running average versus the current middle image of the sequence, iii) motion flow. The first condition requires the first and the last frame of the sequence to have at least partial overlap, the second demands the blurry frame to be related to the sharp frame content-wise. The last condition avoids oscillation (or other failure) cases. We have experimentally found that such a moving average is superior to the constant sum blurring.

## 4 $2MF^2$ Dataset

As it was empirically proved the last few years, the scale of data has a tremendous effect in the performance of systems based on deep convolutional neural networks. Meanwhile, there is still a lack of a large scale database of real-world video data with faces (Ding and Tao (2017)).

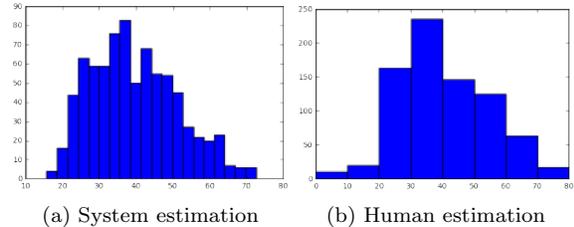

(a) System estimation    (b) Human estimation

Fig. 5: Histograms corresponding to the age estimation of the identities in $2MF^2$. 5a) Estimation by DEX (each bin corresponds to ∼4 years), 5b) estimation by four annotators (each bin corresponds to a decade). Note the similarities of the two histograms; they both indicate that the majority of the identities belong in the age group of $30 - 40$ years old, while they both demonstrate that there are several people from all age groups.

Various databases have been previously published, however all of them have restrictions that jeopardize the pair generation we require. Youtube Faces (YTF) database Wolf et al (2011) includes $3, 425$ videos, several of which are of low resolution and low length (very few frames) and with restricted movement. UMD Faces of Bansal et al (2017) and IJC-B of Whitelam et al (2017) include several thousands of videos, however they do not include sequential frames, which does not enable us to use them for averaging sequential frames. Chung and Zisserman (2016) introduce a dataset for lip-reading, however apart from the constrained conditions of the videos (BBC studios), all the clips include the same duration and a single word is mentioned per clip. Shen et al (2015) introduce 300VW for landmark tracking, however this include very few videos (identities) for our case.

The requirement for high resolution and high fps videos led us to create $2MF^2$ dataset. The dataset was created by downloading public videos from YouTube. Some sample images can be viewed in Fig. 4, while an accompanying video depicting some videos along with the popular 68 shape mark-up can be found in `https://youtu.be/Mz0918XdDew`. Since the original version in Chrysos and Zafeiriou (2017b), we have significantly extended the database, both in number of videos and in number of identities. We have made an effort to create a database with large variations, e.g. to include people with different ages, ethnic groups. The database now includes 10 million frames from 4,000 videos of 850 different identities. The different identities were manually annotated by two different people. On average each identity appears in 11,760 frames. Each person appears in multiple videos, which enables us to capture a wide variation of expressions and external conditions, e.g. illumination, background.

| Dataset name | # Reference | # of videos | # of identities | # of frames | Seq. frames | Identity info |
|---|---|---|---|---|---|---|
| YTF | Wolf et al (2011) | 3,425 | 1,595 | 62,095 | ✓ | ✓ |
| 300VW | Shen et al (2015) | 114 | ∼100 | 218,595 | ✓ | x |
| IJC-B | Whitelam et al (2017) | 7,011 | 1,845 | 55,026 | x | ✓ |
| UMD-Faces V. | Bansal et al (2017) | 22,075 | 3,107 | 3,735,476 | x | ✓ |
| BBC Lip-reading | Chung and Zisserman (2016) | 538,496 | ∼1,000 | 15,616,384 | ✓ | x |
| $2MF^2$ | (this work) | 4,000 | 850 | 10,101,000 | ✓ | ✓ |

Table 2: A comparison of $2MF^2$ to other benchmarks with facial videos in unconstrained conditions. The column 'Seq. frames' denotes whether those videos consist of sequential frames or randomly sampled ones from a video; the column 'Identity info' refers to the meta-data available for identifying different identities.

We have collected annotations for the age of each unique identity by i) human annotators, ii) utilizing the widely-used DEX of Rothe et al (2016). Four human annotators were asked to estimate the age of each unique identity based on the first frame of the video; eight non-overlapping options were available, i.e. $1-10$, $11-20$, etc. The first frame of the video was additionally fed into the popular DEX that estimates the real age of the identity. The automatically derived ages were separated into 20 bins (each bin approximately corresponds to 4 years). The resulting plots for both cases are visualized in Fig.5. The histograms demonstrate that $2MF^2$ includes several samples from each age group[5].

## 5 Experiments

We systematically scrutinized the performance of our proposed method, both internally evaluating the proposed adaptations and against the majority of the publicly available implementations for deblurring. Not only we utilized SSIM as standard metric for image quality, but we also considered deburring as a proxy task and evauated the performance on higher-level tasks in face analysis. Initially, we introduce the implementation and experimental setup and sequentially summarize the experimental results.

### 5.1 Implementation details

**Architecture details**: An off-the-shelf implementation of the HG network as described in Newell et al (2016) is used. The generators $G_{bl}$ and $G_s$ share the same architecture, apart from the first layer that $G_{bl}$ accepts a 6-channel input image, i.e. the concatenation of the blurry image and the output of the HG network. Each generator includes an encoder and a decoder along with skip connections. Both the encoder and the decoder are composed by 8 layers; each convolutional layer is followed by a RELU and batch normalization (Ioffe and Szegedy (2015)). The discriminator consists of 5 layers, while the input images in all sub-networks have $256 \times 256$ shape.

A number of skip connections are added in the generators. Those consist of residual connections (He et al (2016)) and U-net style connections (Ronneberger et al (2015)). The residual connections are added after the first, the third and the fifth layer of the encoder and skip two layers each; the intuition is to propagate the lower level features explicitly without filtering. In the decoder, two similar residual connections are added after the first and the third layers of the decoder. Two U-net style skip connections from the encoder to the decoder are added. Those skip connections enforce the network to learn the residual between the features corresponding to the blurred and the sharp images. This has the impact of faster convergence as empirically detected.

**Training data**: The HG network was trained by synthetically blurred images while the cGAN with the simulated motion blur developed in sec. 3.3. Three million images from MS-Celeb of Guo et al (2016) were blurred using random camera-shake kernels from Hradiš et al (2015). The HG was trained to convergence and then the weights were kept as constants in the training of the cGAN. Conversely, the videos of i) $2MF^2$, ii) BBC trainset of Chung and Zisserman (2016) (sub-sampled to keep every $10^{th}$ video) were utilized to generate the simulated motion blur for cGAN training; 300,000 pairs from $2MF^2$ and 60,000 pairs from BBC.

---

[5] Few samples exist from the children less than 10 and the people older than 70 years old; that is because the videos including such groups are scarce.

Table 3: Details of the conditional Generative Adversarial Network employed. 'Filter size' denotes the size of the filters for the convolutions; the last number denotes the number of output filters. BN stands for batch normalization. Conv denotes a convolutional layer. F-Conv denotes a transposed convolutional layer with fractional-stride. In the conditional GAN, dropout denotes the probability that this layer is kept.

(a) Encoder

| Layer | Filter Size | Stride | BN |
|---|---|---|---|
| Conv. 1 | $4 \times 4 \times 64$ | 1 | × |
| Conv. 2 | $4 \times 4 \times 128$ | 2 | ✓ |
| Conv. 3 | $4 \times 4 \times 256$ | 2 | ✓ |
| Conv. 4 | $4 \times 4 \times 512$ | 2 | ✓ |
| Conv. 5 | $4 \times 4 \times 512$ | 2 | ✓ |
| Conv. 6 | $4 \times 4 \times 512$ | 2 | ✓ |
| Conv. 7 | $4 \times 4 \times 512$ | 2 | ✓ |
| Conv. 8 | $1 \times 1 \times 256$ | 2 | ✓ |

(b) Decoder

| Layer | Filter Size | Stride | BN | Dropout |
|---|---|---|---|---|
| F-Conv. 1 | $1 \times 1 \times 1024$ | 2 | ✓ | 0.5 |
| F-Conv. 2 | $4 \times 4 \times 512$ | 2 | ✓ | 0.5 |
| F-Conv. 3 | $4 \times 4 \times 512$ | 2 | ✓ | 0.5 |
| F-Conv. 4 | $4 \times 4 \times 512$ | 2 | ✓ | 1.0 |
| F-Conv. 5 | $4 \times 4 \times 256$ | 2 | ✓ | 1.0 |
| F-Conv. 6 | $4 \times 4 \times 128$ | 2 | ✓ | 1.0 |
| F-Conv. 7 | $4 \times 4 \times 64$ | 2 | ✓ | 1.0 |
| F-Conv. 8 | $4 \times 4 \times \{1, 3\}$ | 1 | × | 1.0 |

100,000 more pairs were generated by parsing the clips of $2MF^2$ in reverse (temporal) order[6].

**Training details**: The network ran for 60 epochs, trained in a single GPU. We tried to minimize the hyper-parameter tuning by setting $\lambda_{ci} = \lambda_{cg} = \lambda_r$. The values were set experimentally as $\lambda_r = 100$, $\lambda_p = 8$.

### 5.2 Experimental setup

We have included experiments in two popular tasks in face analysis, i.e. landmark localization and face recognition. The benchmarks used for each task were the following:

– **Landmark localization**: The benchmark of 300 videos in-the-wild (300VW) of Shen et al (2015); Chrysos et al (2015) was utilized; this is currently the most established public benchmark for landmark tracking (Chrysos and Zafeiriou (2017a)). This database includes 114 videos; the testset comprises of 64 videos, which are divided into three categories with different degrees of difficulty. Each video is approximately a minute long; each frame includes a single person with 68 markup annotations (Gross et al (2010)) in the face. The 64 videos for testing include 120,000 frames. Such an amount of frames provides a sufficient pool for averaging schemes.

– **Face verification**: The Youtube Faces (YTF) of Wolf et al (2011) includes $3,425$ videos of $1,595$ identities. YTF has been the most popular public benchmark for verification the last few years. Each video includes a single person; there are multiple videos corresponding to the same identity, however the movement in each video is restricted. The length of the videos varies from 50 frames to over $6,000$ with 181 frames as average duration.

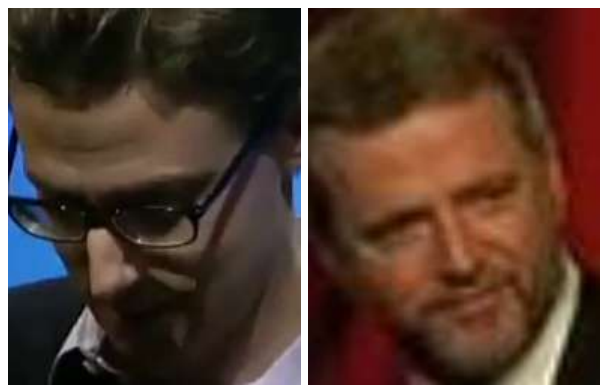

(a) 300VW                    (b) YTF

Fig. 6: Indicative frames from the databases used for testing.

The two datasets include another axis of variation, i.e. facial sizes. The mean size for 300VW is $135 \times 135$, while for YTF is $85 \times 85$. An indicative frame from the datasets is visualized in Fig. 6.

None of the aforementioned databases includes real-world blurred/deblurred pairs, so we have opted to simulate the motion blur with multiple methods mentioned in section 3.3. Specifically, there are three types of blur that we utilize in our experiments:

1. **synthetic blur**: The synthetic blur method of Kupyn et al (2017) was utilized. Random trajectories are generated through a Markov process; the kernel is sampled from the trajectory points and the sharp image is convolved with this kernel.

---

[6] Each reverse blurry frame was compared with the frames exported from the forward pass; the frames that included essentially the same blur were skipped in the reverse order pass.

2. **predefined averaging (PrAvg)**: A predefined number $L$ of frames are added to create the motion blur. The average intensity of the images summed is the blurry image, the ground-truth image is the one in the middle of the interval, i.e. $L/2$. The number $L$ can be visually defined. Even though this is a generic method, it does not take into account neither the intra-class variation, e.g. the movement is not uniform temporally, nor the inter-class variation, e.g. the statistics of each clip are quite different.

3. **Variable length averaging (VLA)**: This is the variable length averaging proposed in section 3.3.

A number of different representative methods have been included in our comparisons. The following energy optimization methods were included: Krishnan et al (2011), Babacan et al (2012); Zhang et al (2013); Xu et al (2013); Pan et al (2014b, 2016). The following state-of-the-art learning methods were compared: Chakrabarti (2016); Kupyn et al (2017); Nah et al (2017). In addition, the domain-specific method of Pan et al (2014a) for face deblurring was included. A considerable computational effort was required to evaluate each of these methods for every single experiment[7], since as reported by Chakrabarti (2016) several of these methods require several minutes per frame.

### 5.3 Self evaluation

To analyze the various components of our method, we trained from scratch different variations of our method. Those variations were: i) one network where the $G_{bl}$ is not conditioned on the HG, i.e. it accepts only the blurry image (called 'identity-noHG'), ii) one network like the previous but trained only on the $2MF^2$ forward data, i.e. 300,000 samples, which was called 'identity-noHG-only2mf2', iii) a network where the was no identity branch, i.e. no $G_s$, which was named 'no-identity', iv) the full version, named 'final'.

To benchmark those variants we used the 300VW and landmark localization. Each blurry/sharp pair was produced by averaging 7 frames, which results in 17,125 test pairs. Each blurry image was deblurred with the four aforementioned variants and then landmark localization was performed in each one of those using the network of Yang et al (2017). Standard quantitative metrics, i.e. AUC, failure rate, were used; the metrics are summarized in the localization experiment in section 5.4.

The quantitative results are summarized in table 4. The following were deduced based on the results: i) the model did benefit from additional labels, ii) the additional conditioning label, i.e. HG network, improved deblurring, iii) the final model with the identity outperformed all the variants.

| Method | PrAvg 7 | | |
|---|---|---|---|
| | *AUC* | *Failure Rate (%)* | *SSIM* |
| identity-noHG-only2mf2 | 0.751 | 8.800 | 0.92031 |
| identity-noHG | 0.759 | 7.895 | 0.88091 |
| no-identity | 0.783 | 5.962 | 0.92883 |
| final | **0.798** | **3.947** | **0.93501** |
| *Nr. of test images* | 17,125 | | |

Table 4: Self evaluation results on 300VW. The experimental outcomes indicate that i) the additional conditioning label did have a positive impact, ii) the identity outperforms the 'no-identity' variant.

### 5.4 Landmark localization

If the deblurred images indeed resemble the statistics of sharp facial images, then a localization method should be close to the localization of the original ground-truth images. To assess this similarity, we utilized the 300VW as a testbed to compare the localization of the deblurred images. The frames of 300VW were blurred and each comparison method was used to deblur them. Sequentially, the state-of-the-art landmark localization method of Yang et al (2017), winner of the Menpo Challenge of Zafeiriou et al (2017), was used to perform landmark localization in the deblurred images. Apart from the comparison methods, an 'oracle' was implemented. The 'oracle' represents the perfect deblurring method, i.e. the deblurred images are identical to the latent sharp images. We used the oracle to indicate the upper bound of the performance of the deblurring methods.

The following error metrics are reported for this experiment:

- cumulative error distribution (CED) plot: It depicts the percentage of images (y-axis) that have up to a certain percentage of error (x-axis).
- area under the curve (AUC): A scalar that is the area under the CED plot.
- failure rate: The localization error is cut-off at 0.08; any larger error is considered a failure to localize the fiducial points. Failure rate is the percentage of images with error larger than the cut-off error.

---

[7] The public implementations of all methods were used.

- structural similarity (SSIM) by Wang et al (2004): Image quality metric typically used to report the quality of the deblurred images.
- cosine distance distribution plot: The embedding of the face is extracted per frame with faceNet of Schroff et al (2015); similarly the embedding of the sharp image is extracted and the cosine distance of the two is computed. A cumulative distribution of those cosine distances is plotted.

The CED plot, AUC and Failure rate consist standard localization error metrics; we utilize the same conventions as in Chrysos et al (2017), i.e. the error is the mean euclidean distance of the points, normalized by the diagonal of the ground-truth bounding box.

The following schemes for simulating blur were utilized: i) predefined averaging with $L \in \{7, 9, 11, 15, 21\}$, ii) VLA, iii) synthetic blur in ground-truth images of VLA. The different options of predefined averaging considered allowed us to assess the robustness under mild differences in blurring averages.

The quantitative results for the predefined averaging cases are depicted in tables 5, 6; the CED plots[8] of the top three performing methods (based on the AUC) are visualized in Fig. 9, 10. The complete CED plots are visualized in the supplementary material. The reported metrics validate that optimization methods did not perform as well as learning-based methods; the only method that consistently performed well was Pan et al (2016). Conversely, the learning-based methods improved the results of the blurred, while our method consistently outperformed the rest in all cases. As it is noticeable from the CED plots, in the region of small errors, i.e. $< 0.02$, a large number of averaged frames deteriorated the fitting considerably. Additionally, the majority of the methods were robust to small changes in the number of averaged frames, i.e. from 7 to 9 or from 9 to 11. On the contrary, the difference between averaging 7 and 21 frames was noticeable in most methods; in our case the decrease in the performance was less than the compared methods.

Similar conclusions hold for the experiments with the VLA and the synthetic kernels schemes, the results of which exist in tables 6 and 7. In both cases, our method increased the margin from the comparison methods. That is attributed to the very diverse number of blurs that our method was trained on. On the contrary, the prior art of Nah et al (2017) suffered for slightly modified conditions than the predefined averaging.

In all seven cases with different blurs examined, we verified that our method was robust to mediocre and severe blurs. In Fig. 7, 8 two images with the landmarks overlaid are visualized, while in Fig. 13, 14, 15, 16 and 17 qualitative images with different cases are illustrated[9].

Aside of the state-of-the-art network of Yang et al (2017), we selected the top three performing methods and repeated two experiments with an alternative localization method. The method chosen was CFSS of Zhu et al (2015), which was the top performing regression-based method. The results, which are in the supplementary material, ranked our method's deblurred images as the top performing ones.

Apart from the localization of the fiducial points, we wanted to assess the preservation of the identity in the various deblurring methods. Since there is no ground-truth information for the identity encoding, we opt to report a soft-metric instead. The embeddings of the widely used model of faceNet were adopted; we measured the cosine distance between each deblurred frame's embedding and the respective ground-truth's embedding. In the perfect identity preservation case, the cosine distance distribution plot should be a dirac delta around one; a narrow distribution centered to one denotes proximity to the embeddings of the ground-truth face. The results in Fig. 11, 12 indicate that our method is robust in different cases of blur and has a cosine distance distribution closer to the ground-truth than the compared methods.

### 5.5 Face verification

We utilized the Youtube Faces (YTF) dataset for performing face verification. The video frames were averaged to generate the blurry/sharp pairs; each deblurring method was applied and then the deblurred outputs were used for face verification. Assessing the accuracy of each method would directly allow us to compare which method results in facial representations that maintain the identity.

The complete setup for the experiment was the following: The verification was performed in the deblurred images by extracting representation from faceNet of Schroff et al (2015). We employed the predefined averaging with i) 7, ii) 11 frames[10]. The error metric used was the mean accuracy along with the computed standard deviation. The results are summarized in table 8. The learning-based methods performed prefer-

---

[8] The results of the averaging 7 or 9 frames are similar, hence the related CED plot is omitted.

[9] A supplementary video with deblurring examples is uploaded on https://youtu.be/Zb_6taCOOi4.

[10] We did not utilize the VLA method, since some of the videos were shorter than those required for the rolling averaging of VLA.

| Method | PrAvg 7 | | | PrAvg 9 | | | PrAvg 11 | | |
|---|---|---|---|---|---|---|---|---|---|
| | AUC | Failure Rate (%) | SSIM | AUC | Failure Rate (%) | SSIM | AUC | Failure Rate (%) | SSIM |
| Krishnan et al (2011) | 0.706 | 11.013 | 0.7857 | 0.734 | 5.801 | 0.78183 | 0.728 | 6.294 | 0.7909 |
| Babacan et al (2012) | 0.730 | 9.950 | 0.92877 | 0.735 | 7.796 | 0.91614 | 0.728 | 7.993 | 0.91913 |
| Zhang et al (2013) | 0.692 | 11.276 | 0.79779 | 0.713 | 6.956 | 0.79298 | 0.711 | 7.129 | 0.80369 |
| Xu et al (2013) | 0.451 | 42.377 | 0.86214 | 0.348 | 55.125 | 0.85477 | 0.451 | 41.528 | 0.85988 |
| Pan et al (2014b) | 0.709 | 7.527 | 0.74978 | 0.699 | 8.194 | 0.74378 | 0.692 | 8.434 | 0.75232 |
| Pan et al (2014a) | 0.745 | 5.255 | 0.80171 | 0.734 | 6.077 | 0.80548 | 0.727 | 6.304 | 0.79982 |
| Pan et al (2016) | **0.768** | **4.502** | **0.86626** | **0.756** | **4.968** | **0.85379** | **0.749** | **5.440** | **0.86143** |
| Chakrabarti (2016) | 0.504 | 37.080 | 0.88817 | 0.480 | 39.708 | 0.88462 | 0.466 | 41.403 | 0.88281 |
| Nah et al (2017) | **0.792** | **3.235** | **0.94294** | **0.779** | **3.837** | **0.93318** | **0.771** | **4.471** | **0.92504** |
| Kupyn et al (2017) | 0.766 | 5.647 | 0.88605 | 0.749 | 6.902 | 0.8946 | 0.739 | 7.503 | 0.89004 |
| ours | **0.798** | **3.947** | **0.94501** | **0.788** | **4.525** | **0.93854** | **0.781** | **5.018** | **0.93221** |
| blurred | 0.743 | 8.876 | 0.94078 | 0.747 | 6.680 | 0.92705 | 0.739 | 7.043 | 0.92905 |
| oracle | 0.827 | 1.816 | - | 0.829 | 1.605 | - | 0.829 | 1.612 | - |
| Nr. of test images | 17,125 | | | 13,083 | | | 10,422 | | |

Colouring denotes the methods' performance ranking per category (based on AUC): ■ first ■ second ■ third

Table 5: Quantitative results on the experiment on landmark localization of sec. 5.4. The first row dictates the blurring process. In this plot from left to right predefined averaging with (a) 7, (b) 9, (c) 11 frames.

| Method | PrAvg 15 | | | PrAvg 21 | | | VLA | | |
|---|---|---|---|---|---|---|---|---|---|
| | AUC | Failure Rate (%) | SSIM | AUC | Failure Rate (%) | SSIM | AUC | Failure Rate (%) | SSIM |
| Krishnan et al (2011) | 0.719 | 6.717 | 0.78849 | 0.707 | 7.094 | 0.78858 | 0.731 | 4.192 | 0.78706 |
| Babacan et al (2012) | 0.710 | 8.807 | 0.90831 | 0.684 | 10.641 | 0.89621 | 0.715 | 6.817 | 0.89318 |
| Zhang et al (2013) | 0.705 | 7.165 | 0.80162 | 0.689 | 8.122 | 0.79893 | 0.721 | 4.879 | 0.80180 |
| Xu et al (2013) | 0.446 | 41.552 | 0.85767 | 0.438 | 41.435 | 0.85059 | 0.390 | 48.459 | 0.84882 |
| Pan et al (2014b) | 0.681 | 8.536 | 0.74793 | 0.665 | 9.170 | 0.74447 | 0.692 | 6.852 | 0.75602 |
| Pan et al (2014a) | 0.716 | 6.609 | 0.79162 | 0.694 | 7.960 | 0.78256 | 0.727 | 4.280 | 0.78333 |
| Pan et al (2016) | **0.735** | **5.808** | **0.85166** | **0.714** | **6.590** | **0.8409** | **0.745** | **3.734** | **0.84249** |
| Chakrabarti (2016) | 0.437 | 44.375 | 0.87871 | 0.405 | 48.065 | 0.87234 | 0.410 | 45.006 | 0.84582 |
| Nah et al (2017) | **0.754** | **5.225** | **0.91227** | **0.730** | **7.255** | **0.89832** | **0.767** | **3.012** | **0.90223** |
| Kupyn et al (2017) | 0.720 | 8.386 | 0.86222 | 0.694 | 9.734 | 0.85297 | 0.728 | 5.513 | 0.85285 |
| ours | **0.769** | **5.482** | **0.91316** | **0.748** | **6.751** | **0.90257** | **0.786** | **3.118** | **0.9075** |
| blurred | 0.721 | 7.667 | 0.91745 | 0.695 | 9.270 | 0.90388 | 0.725 | 5.866 | 0.90223 |
| oracle | 0.832 | 1.276 | - | 0.833 | 1.411 | - | 0.827 | 1.568 | - |
| Nr. of test images | 7,369 | | | 4,962 | | | 5,677 | | |

Colouring denotes the methods' performance ranking per category (based on AUC): ■ first ■ second ■ third

Table 6: Second part of the quantitative results for the landmark localization experiment of sec. 5.4.

ably to the optimization-based, while our method outperformed all the rest.

### 5.6 Real-world blurry video

A video with extreme motion blur was captured with a high-precision camera; ground-truth frames are not available, we only have the 160 blurry frames. To allow a quantitative comparison, a sharp frame of the video was selected for extracting the identity embeddings with faceNet. Then each method deblurred the images, the embeddings were extracted and compared as in the localization experiment (section 5.4). The respective cosine distance distribution plot is depicted in Fig. 18. As it is noticeable in the plot our method is ranked as the one closest to the identity embedding from the sharp frame. An indicative frame is plotted in Fig. 19.

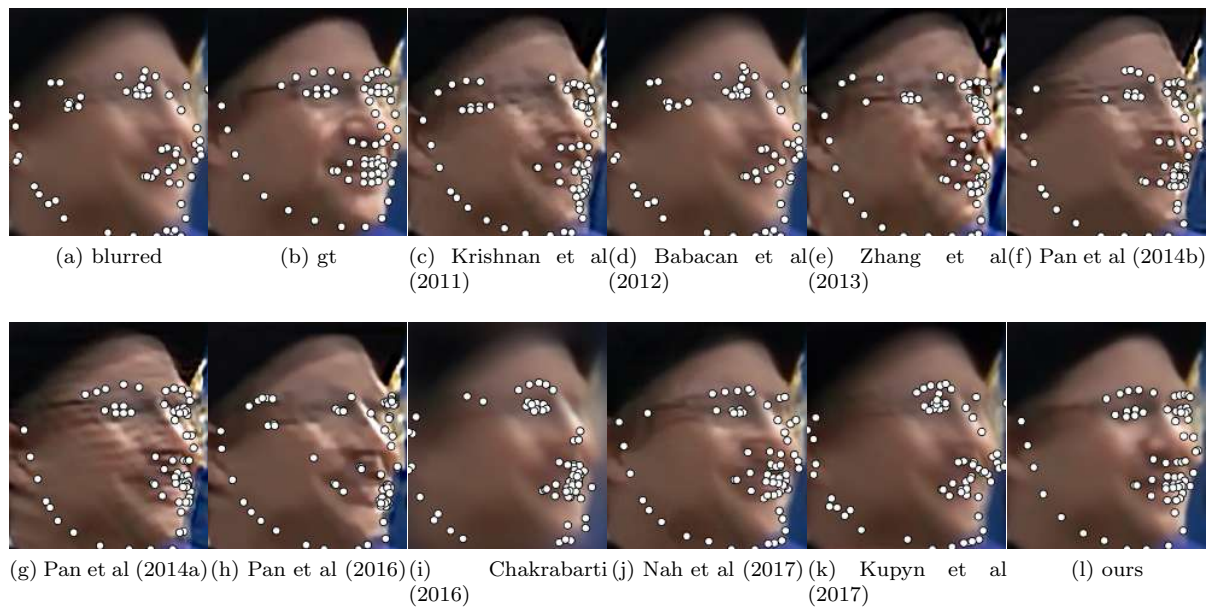

(a) blurred  (b) gt  (c) Krishnan et al (2011)  (d) Babacan et al (2012)  (e) Zhang et al (2013)  (f) Pan et al (2014b)

(g) Pan et al (2014a)  (h) Pan et al (2016)  (i) Chakrabarti (2016)  (j) Nah et al (2017)  (k) Kupyn et al (2017)  (l) ours

Fig. 7: (Preferably viewed in color) Landmarks overlaid in the deblurred images as described in section 5.4.

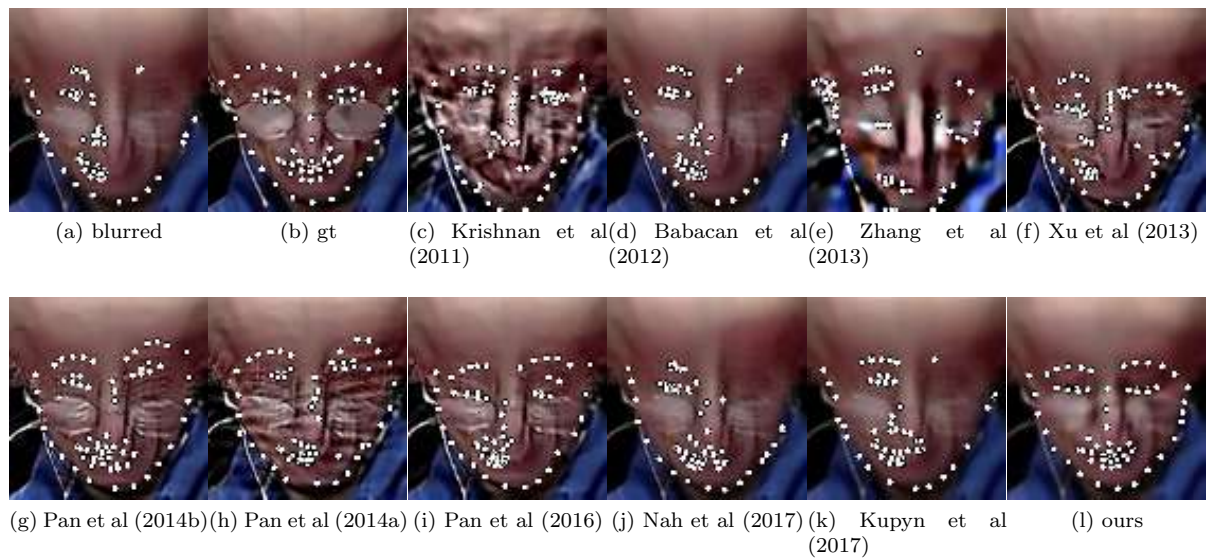

(a) blurred  (b) gt  (c) Krishnan et al (2011)  (d) Babacan et al (2012)  (e) Zhang et al (2013)  (f) Xu et al (2013)

(g) Pan et al (2014b)  (h) Pan et al (2014a)  (i) Pan et al (2016)  (j) Nah et al (2017)  (k) Kupyn et al (2017)  (l) ours

Fig. 8: (Preferably viewed in color) Landmarks overlaid in the deblurred images as described in section 5.4.

## 6 Conclusion

In this work, we introduce a method for performing motion deblurring of faces. Our method is based on the preservation of the identity of the face. We learn a two-step system, where in the first step a strong discriminative network restores the low-frequency details; in the second step the high-frequency details are re-stored. To train this system, we devise a new way of simulating motion blur by averaging a variable number of frames. The frames originate from videos in the $2MF^2$ dataset that we collect for this task. We test our system in a thorough experimentation, using both quality metrics typically used in deblurring but also more established quantitative tasks in face analysis, i.e. landmark localization and face verification. In both tasks and in all the conducted experiments our method performs favourably to the compared methods setting the new state-of-the-art for face deblurring.

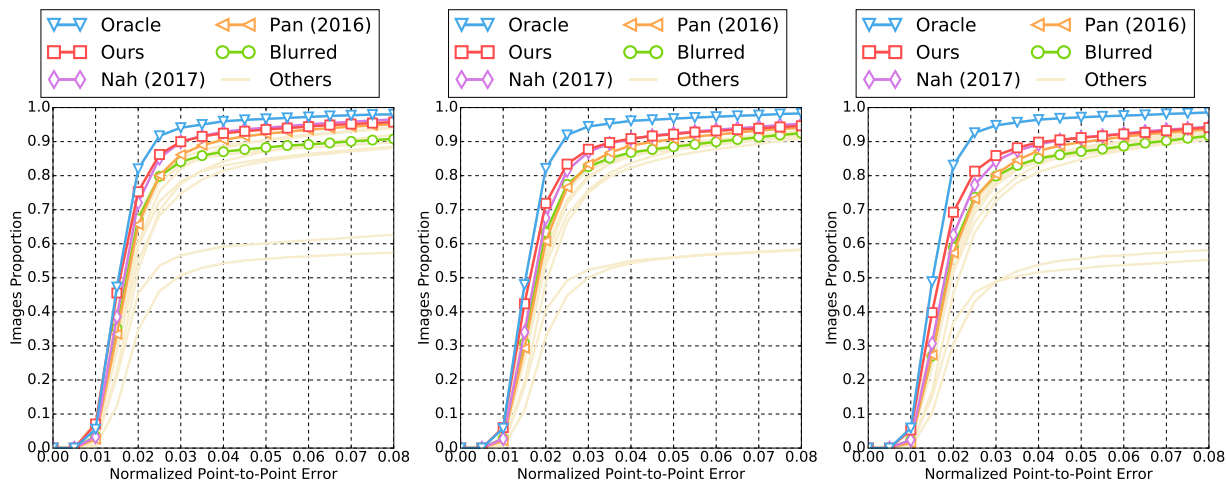

Fig. 9: CED plots for the landmark localization experiment of sec. 5.4. To avoid cluttering the plot, only the top 3 methods along with the oracle and the original blur performance are plotted. From left to right the plots correspond to the experiments with blurring process: predefined averaging with (a) 7, (b) 11, (c) 15 frames.

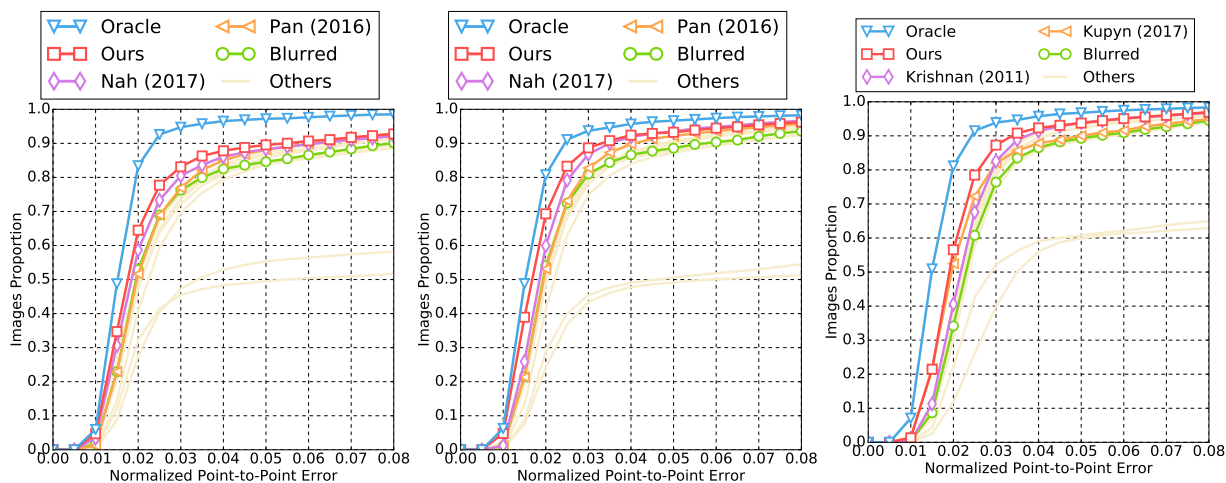

Fig. 10: Continuation of the CED plots of the landmark localization experiment. From left to right the plots correspond to the experiments with blurring process: (a) predefined averaging with 21 frames, (b) VLA, (c) synthetic blur.

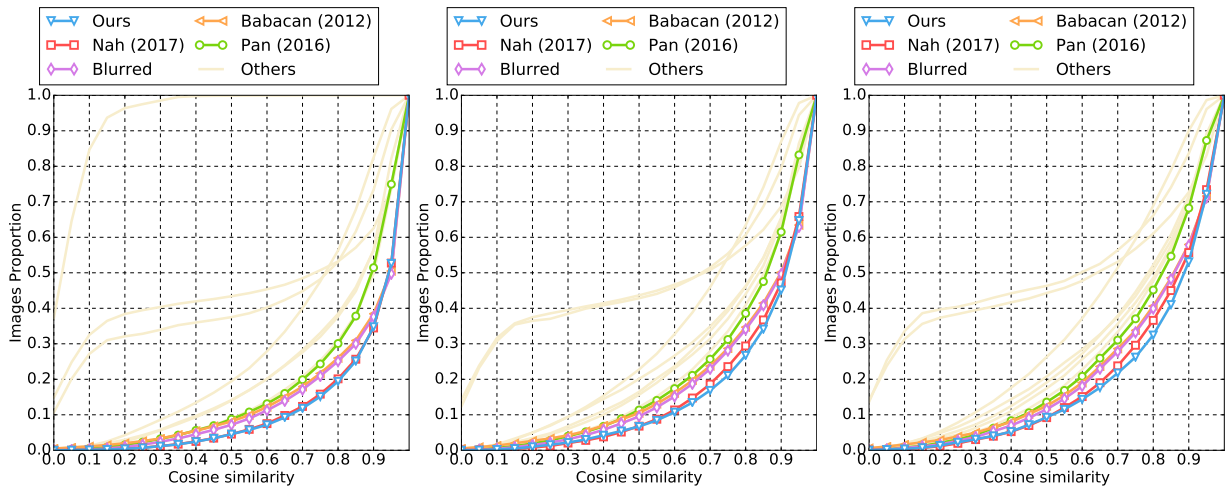

Fig. 11: Cosine distance distribution plots for the landmark localization experiment of sec. 5.4. To avoid cluttering the plot, only the top four methods along with the original blur performance are plotted. The narrower distributions concentrated around one declare closer representation of the ground-truth identity. Please find further details in the text. From left to right the plots correspond to the experiments with blurring process: predefined averaging with (a) 7, (b) 11, (c) 15 frames.

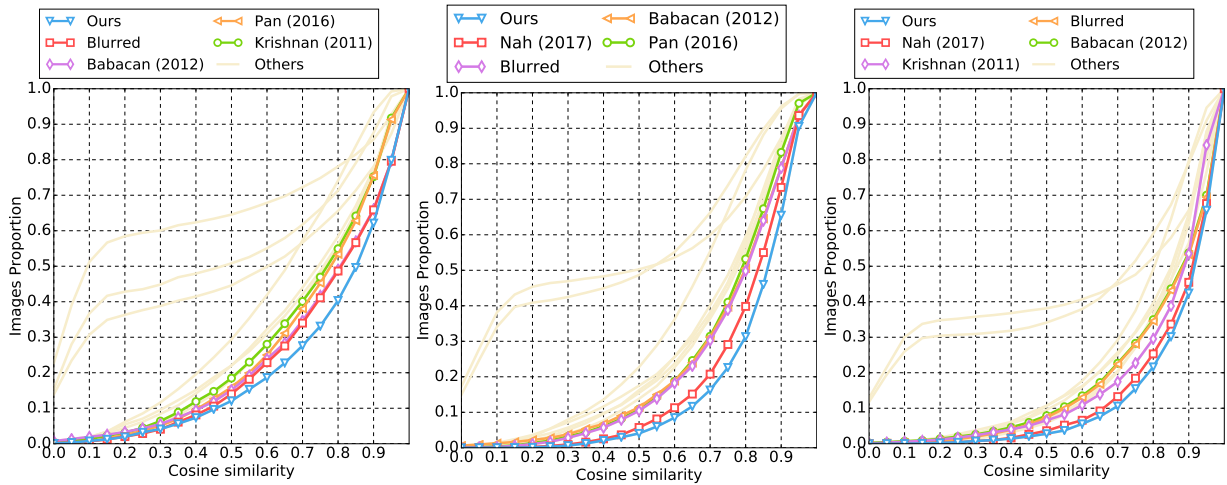

Fig. 12: Continuation of the cosine distance distribution plots from Fig. 11. From left to right the plots correspond to the experiments with blurring process: (a) predefined averaging with 21 frames, (b) VLA, (c) synthetic blur.

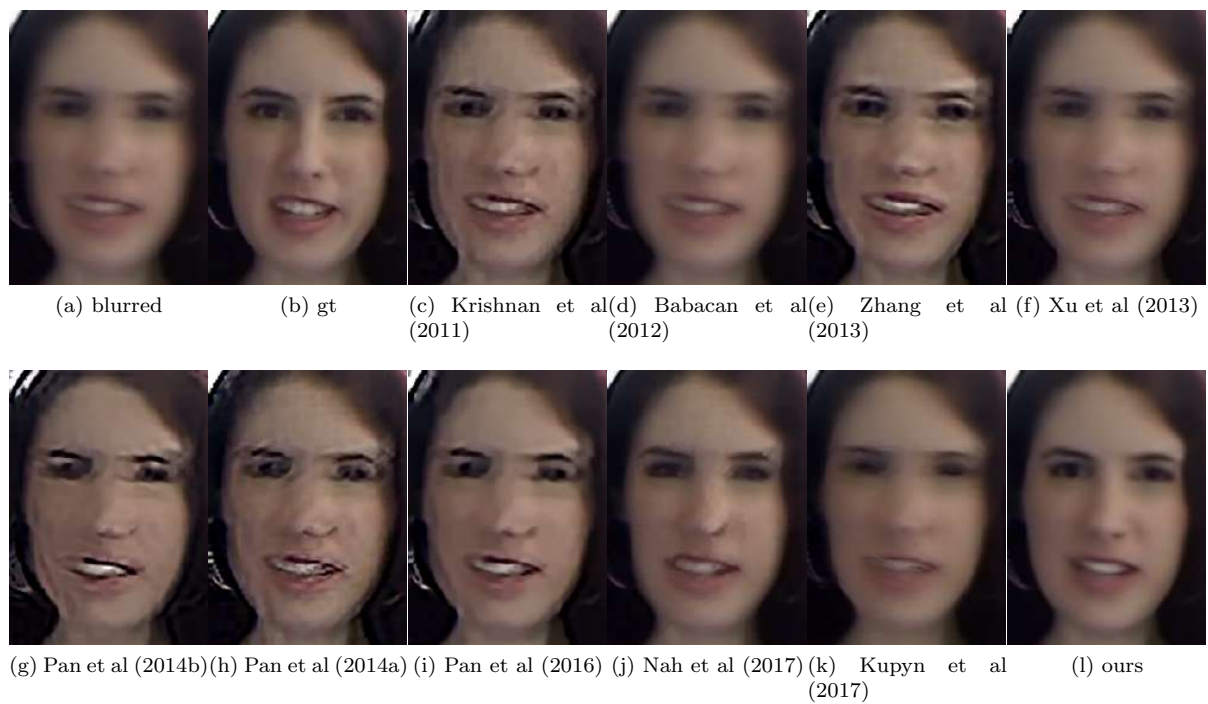

(a) blurred     (b) gt     (c) Krishnan et al (2011)   (d) Babacan et al (2012)   (e) Zhang et al (2013)   (f) Xu et al (2013)

(g) Pan et al (2014b) (h) Pan et al (2014a) (i) Pan et al (2016) (j) Nah et al (2017) (k) Kupyn et al (2017)     (l) ours

Fig. 13: Visual results. The majority of the existing methods fail to deblur the eyes and the nose; even the state-of-the-art method of Nah et al (2017) does not manage to yield a realistic face. On the contrary, our method outputs a realistic face with both the eyes and the nose accurately deblurred.

| Method | Synthetic | | |
|---|---|---|---|
| | *AUC* | *Failure Rate (%)* | *SSIM* |
| Krishnan et al (2011) | 0.736 | 2.774 | 0.76720 |
| Babacan et al (2012) | 0.686 | 5.344 | 0.83956 |
| Zhang et al (2013) | 0.730 | 2.850 | 0.77321 |
| Xu et al (2013) | 0.437 | 36.845 | 0.76055 |
| Pan et al (2014b) | 0.715 | 3.486 | 0.74478 |
| Pan et al (2014a) | 0.718 | 3.639 | 0.72402 |
| Pan et al (2016) | 0.727 | 2.723 | 0.77778 |
| Chakrabarti (2016) | 0.475 | 34.631 | 0.84699 |
| Nah et al (2017) | 0.729 | 2.774 | 0.83791 |
| Kupyn et al (2017) | 0.733 | 4.580 | 0.84477 |
| ours | **0.764** | **2.672** | **0.87141** |
| blurred | 0.697 | 4.656 | 0.85225 |
| oracle | 0.830 | 1.425 | - |
| *Nr. of test images* | 3,930 | | |

Ranking (based on AUC): ■ first ■ second ■ third

Table 7: Third (last) part of the quantitative results for the landmark localization experiment of sec. 5.4.

| Method | *PrAvg 7* | *PrAvg 11* |
|---|---|---|
| | *Mean acc. ± std* | *Mean acc. ± std* |
| Krishnan et al (2011) | 0.7344 ± 0.0155 | 0.7146 ± 0.0203 |
| Babacan et al (2012) | 0.8194 ± 0.0193 | **0.8024 ± 0.0168** |
| Zhang et al (2013) | 0.7002 ± 0.0219 | 0.6964 ± 0.0150 |
| Xu et al (2013) | 0.5346 ± 0.0264 | 0.5348 ± 0.0176 |
| Pan et al (2014b) | 0.6768 ± 0.0234 | 0.6654 ± 0.0213 |
| Pan et al (2014a) | 0.7386 ± 0.0228 | 0.7190 ± 0.0145 |
| Pan et al (2016) | 0.7648 ± 0.0241 | 0.7508 ± 0.0217 |
| Chakrabarti (2016) | 0.5510 ± 0.0272 | 0.5410 ± 0.0206 |
| Nah et al (2017) | **0.8434 ± 0.0188** | **0.8106 ± 0.0156** |
| Kupyn et al (2017) | **0.8244 ± 0.0161** | 0.7942 ± 0.0163 |
| ours | **0.8482 ± 0.0179** | **0.8212 ± 0.0087** |
| blurred | 0.8414 ± 0.0206 | 0.8154 ± 0.0136 |
| gt | 0.9006 ± 0.0118 | 0.9044 ± 0.0134 |

Ranking (based on mean accuracy): ■ first ■ second ■ third

Table 8: Quantitative results in the face verification experiment of section 5.5.

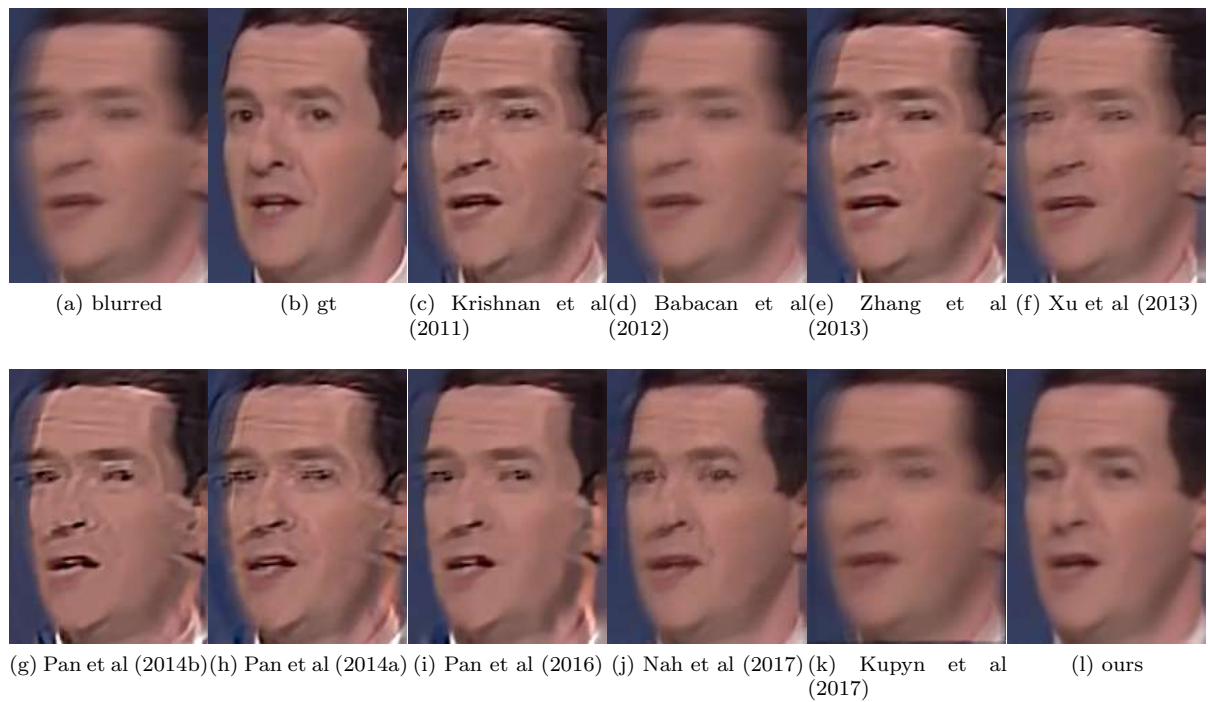

(a) blurred              (b) gt              (c)  Krishnan  et  al (d)  Babacan  et  al (e)  Zhang   et   al (f) Xu et al (2013)
                                             (2011)              (2012)              (2013)

(g) Pan et al (2014b) (h) Pan et al (2014a) (i) Pan et al (2016) (j) Nah et al (2017) (k)  Kupyn   et   al (l) ours
                                                                                     (2017)

Fig. 14: The nature of the blur caused ghost artifacts in the outputs of the majority of the methods. Some of them are more subtle, e.g. in Pan et al (2016); Nah et al (2017), however they are visible by zooming-in the figures. Our method avoided such artifacts and returned a plausible face.

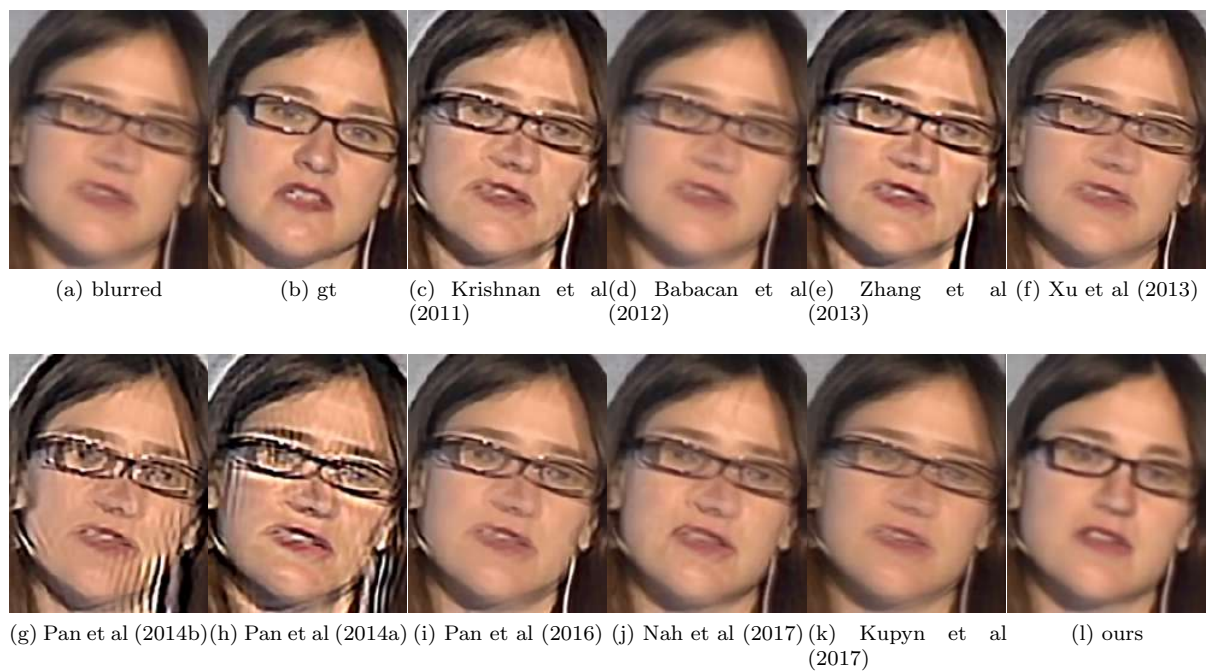

(a) blurred              (b) gt              (c)  Krishnan  et  al (d)  Babacan  et  al (e)  Zhang   et   al (f) Xu et al (2013)
                                             (2011)              (2012)              (2013)

(g) Pan et al (2014b) (h) Pan et al (2014a) (i) Pan et al (2016) (j) Nah et al (2017) (k)  Kupyn   et   al (l) ours
                                                                                     (2017)

Fig. 15: The glasses are severely affected by the motion blur in this case; the compared method, even those for generic deblurring, fail to restore the glasses and the nose region, while our method returns a plausible outcome.

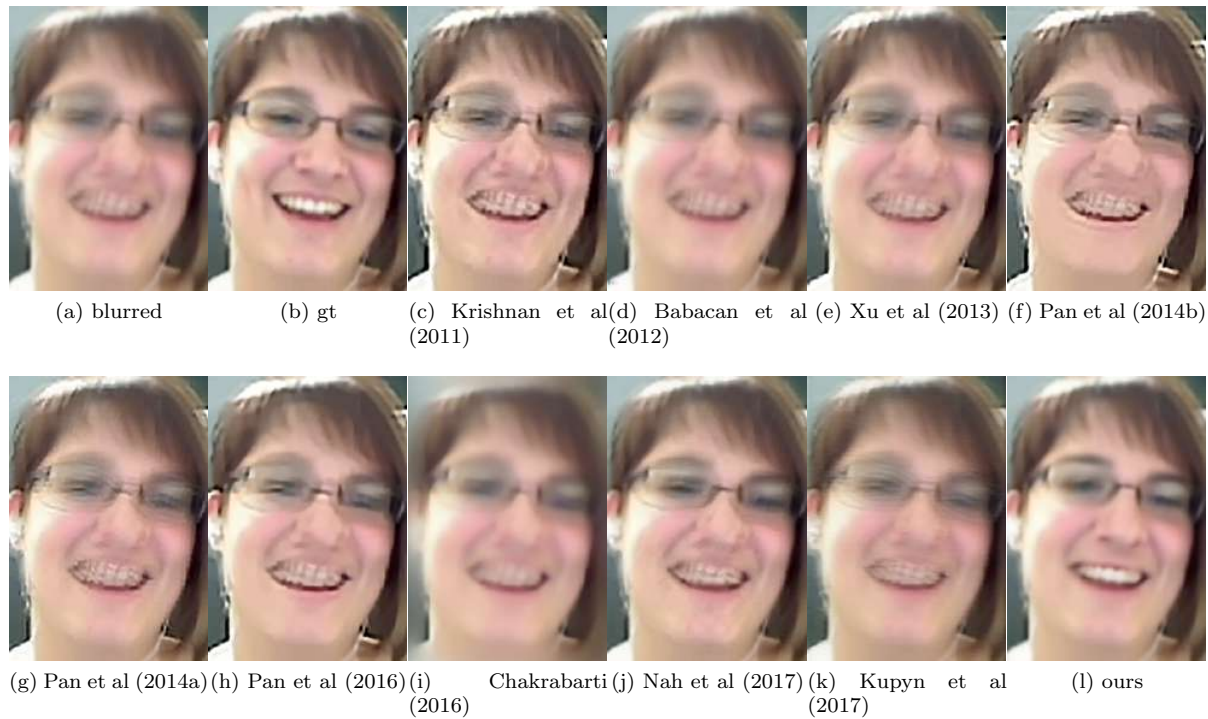

(a) blurred      (b) gt      (c) Krishnan et al (2011)    (d) Babacan et al (2012)    (e) Xu et al (2013)    (f) Pan et al (2014b)

(g) Pan et al (2014a)    (h) Pan et al (2016)    (i) Chakrabarti (2016)    (j) Nah et al (2017)    (k) Kupyn et al (2017)    (l) ours

Fig. 16: The movement of the face caused severe blur in the mouth region. All the compared methods fail to deblur the mouth, our method does that accurately.

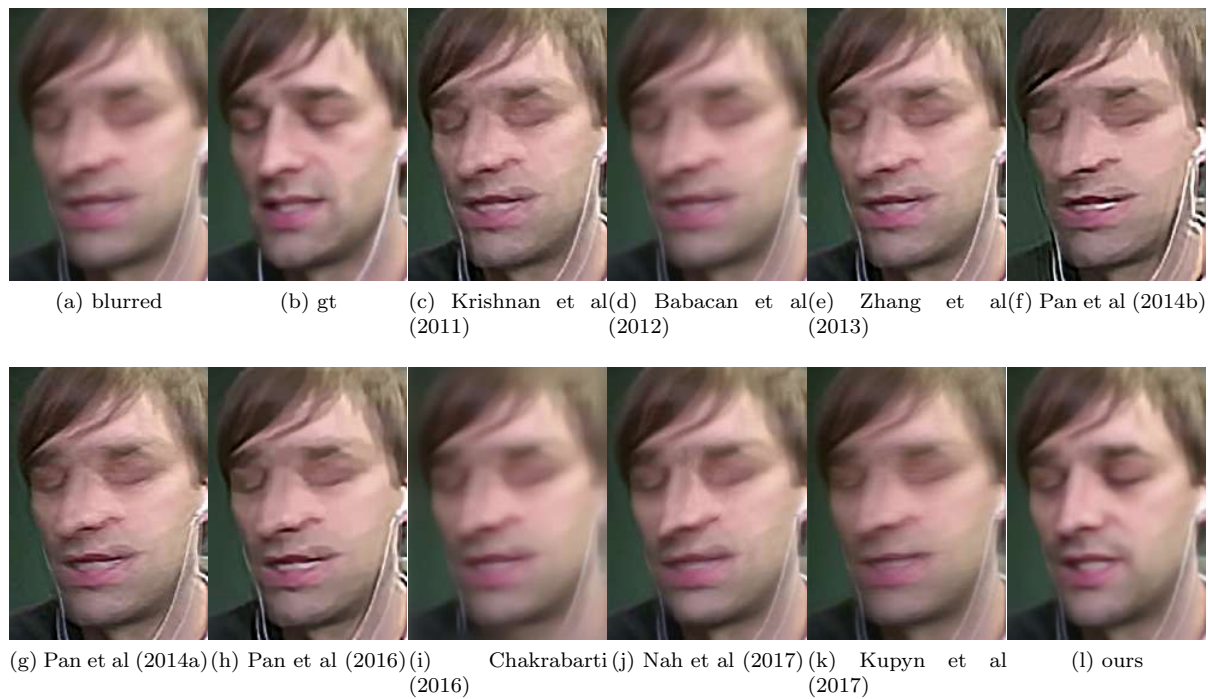

(a) blurred    (b) gt    (c) Krishnan et al (2011)    (d) Babacan et al (2012)    (e) Zhang et al (2013)    (f) Pan et al (2014b)

(g) Pan et al (2014a)    (h) Pan et al (2016)    (i) Chakrabarti (2016)    (j) Nah et al (2017)    (k) Kupyn et al (2017)    (l) ours

Fig. 17: The horizontal rotation caused severe blur in the nose, which our method deblurs successfully in comparison to the rest methods.

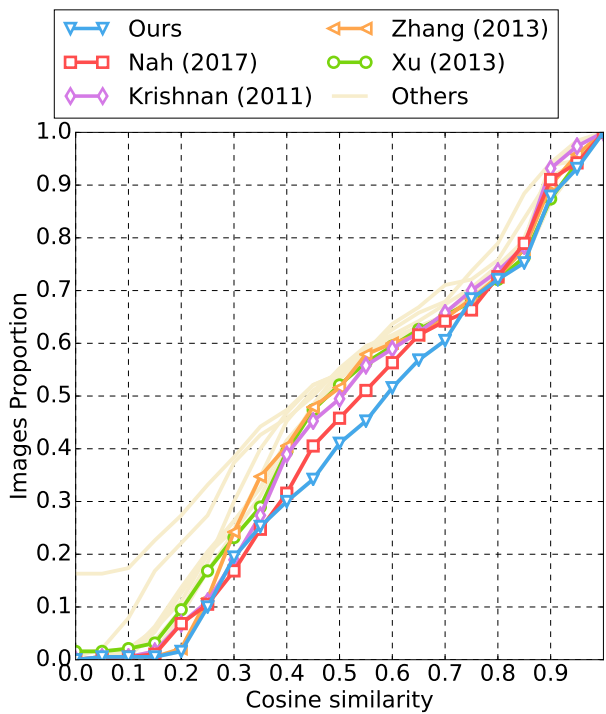

Fig. 18: Cosine distance distribution plots for the real-world blurry video of sec. 5.6. To avoid cluttering the plot, only the top four methods along with the original blur performance are plotted. The narrower distributions concentrated around one declare closer representation of the ground-truth identity. The legends from top to bottom, left to right declare the ranking of the methods.

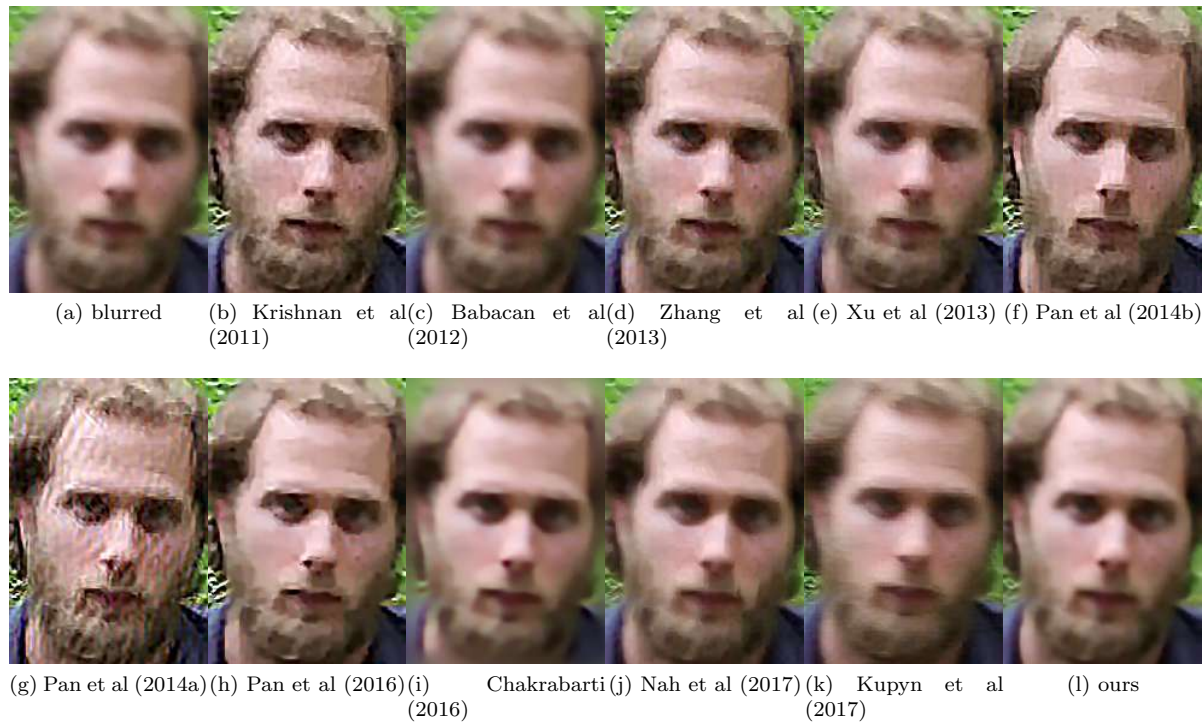

(a) blurred   (b) Krishnan et al (c) Babacan et al (d) Zhang et (e) Xu et al (2013) (f) Pan et al (2014b)
(2011)              (2012)              al (2013)

(g) Pan et al (2014a) (h) Pan et al (2016) (i) Chakrabarti (j) Nah et al (2017) (k) Kupyn et al (l) ours
                                        (2016)                                 (2017)

Fig. 19: Deblurring in a real-world blurry image. Even the state-of-the-art methods of Nah et al (2017); Kupyn et al (2017) over-smooth the blurred image (please zoom-in for further understanding). On the contrary our method yields an improved outcome.